\documentclass{isprs}

\usepackage{subcaption}
\usepackage{setspace}
\usepackage{geometry} 
\usepackage{epstopdf}
\usepackage[labelsep=period]{caption}  
\usepackage[british]{babel} 
\usepackage[hang]{footmisc}

\usepackage[pagebackref,breaklinks,colorlinks,citecolor=cvprblue]{hyperref}

\usepackage[dvipsnames]{xcolor}

\usepackage{adjustbox}

\renewcommand{\vec}[1]{\boldsymbol{#1}}
\newcommand{\matr}[1]{\boldsymbol{#1}}

\newcommand{\dotprod}[2]{\left\langle#1,#2\right\rangle}

\def\censorloopword#1 #2\nil{%
  \phantom{#1} 
  \ifx&#2&
    \let\next\relax
  \else
    \def\next{\censorloopword#2\nil}
  \fi
  \next\ignorespaces}

\usepackage{times}    
\usepackage{xspace}
\usepackage{graphicx}
\usepackage{amsmath}
\usepackage{amssymb}
\usepackage{booktabs}
\usepackage[font=footnotesize,skip=3pt,subrefformat=parens]{subcaption}

\renewcommand{\vec}[1]{\boldsymbol{#1}}

\newcommand{\eg}{\textit{e.g.}, }
\newcommand{\ie}{\textit{i.e.}, }
\newcommand{\etal}{\textit{et al. }}

\newcommand{\className}[1]{\textbf{\textit{#1}}}

\definecolor{cvprblue}{rgb}{0.21,0.49,0.74}
\usepackage[pagebackref,breaklinks,colorlinks,citecolor=cvprblue]{hyperref}

\begin{document}


\title{Weakly-Supervised Learning for Tree Instances Segmentation in Airborne Lidar Point Clouds}

\author{Swann Emilien Céleste Destouches\textsuperscript{1}, Jesse Lahaye\textsuperscript{1}, Laurent V. Jospin \textsuperscript{1}, Jan Skaloud \textsuperscript{1}}

\address{
	\textsuperscript{1 }Environmental Sensing \&  Observation Laboratory (ESO), Ecole Polytechnique Fédérale de Lausanne (EPFL), Switzerland – \\ swann.destouches@alumni.epfl.ch, (jesse.lahaye, laurent.jospin, jan.skaloud)@epfl.ch
}


\abstract{
Tree instance segmentation of airborne laser scanning (ALS) data is of utmost importance for forest monitoring, but remains challenging due to variations in the data caused by factors such as sensor resolution, vegetation state at acquisition time, terrain characteristics, etc. Moreover, obtaining a sufficient amount of precisely labeled data to train fully supervised instance segmentation methods is expensive. To address these challenges, we propose a weakly supervised approach where labels of an initial segmentation 
result obtained either by a non-finetuned model or a closed form algorithm are provided as a quality rating by a human operator. The labels produced during the quality assessment are then used to train a rating model, whose task is to classify a segmentation output into the same classes as specified by the human operator. Finally, the segmentation model is finetuned using feedback from the rating model.
This in turn improves the original segmentation model by 34\% in terms of correctly identified tree instances while considerably reducing the number of non-tree instances predicted. Challenges still remain in data over sparsely forested regions characterized by small trees (less than two meters in height) or within complex surroundings containing shrubs, boulders, etc. which can be confused as trees where the performance of the proposed method is reduced.
}

\maketitle
\sloppy

\section{Introduction}
\label{sec:intro}

Quantifying tree distribution in forests is an important application of 3D vision to either estimate its economic potential (\eg exploitable wood volume) \cite{rs12071078}, to measure the impact of climate change \cite{FOUQUERAY2020117880}, to monitor the impact on slope stability \cite{Jiang2023}, or to estimate potential for carbon capture  \cite{10.1111_2041-210X.12575}.

Individual tree crown segmentation (ITCs) in generic 3D forest data remains a challenging task. While deep learning models have improved it to some extent, they still struggle to deal with variability in data resolution, species, seasonal changes (leaf on/off) and artifacts present in the terrain, especially in mountainous areas where cliffs and boulders can interfere with the segmentation. Despite recent improvements toward generalization \cite{wielgosz_segmentanytree_2024}, fine-tuning remains a necessity which poses a challenge for general applicability of existing methods. Producing manual labels for segmentation tasks is time consuming and error prone, both for semantic \cite{Shimoda_2019_ICCV} and instance \cite{Cheng_2023_CVPR} segmentation.
On the other hand, access to high quality labeled data is required by deep-learning approaches for their fine-tuning on novel target area(s) of interest (\eg, alpine ecotones), which remains an issue \cite{isprs-annals-X-1-2024-67-2024}.

\begin{figure}[t]
    \centering
    \includegraphics[width=\linewidth]{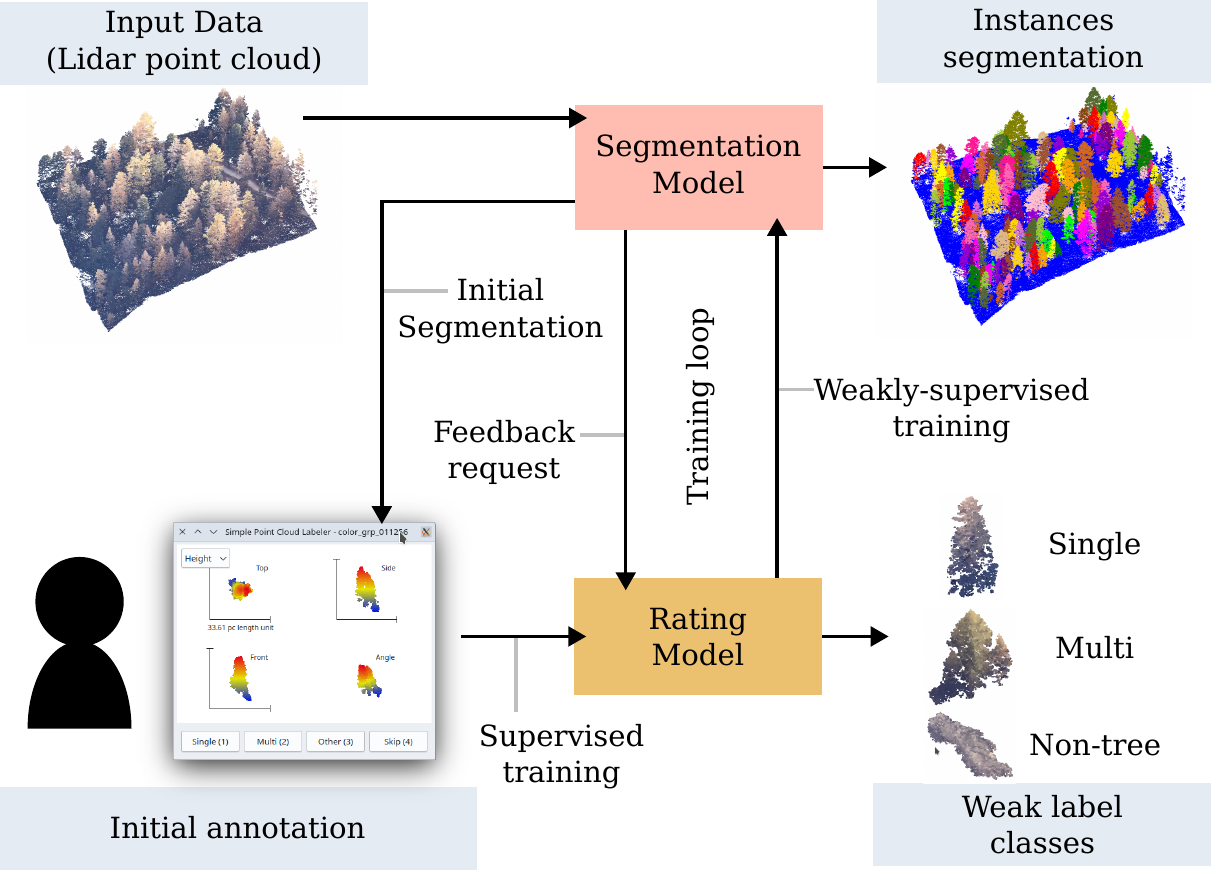}
    \vspace{-15pt}
    \caption{Proposed approach for weakly supervised training of tree segmentation.}
    \label{fig:graphical_abstract}
\end{figure}

To address these challenges, we propose an approach inspired by Reinforcement Learning from Human Feedback (RLHF  \cite{10.1145/3743127}). Instead of labeling the data with a segmentation mask, we provide feedback on samples from the segmentation results, which is much faster to generate by a human operator. Then we train a classification model (rating model) to imitate the rating of the human operator, which in turn provides feedback to the segmentation model; see Figure~\ref{fig:graphical_abstract}. While this approach is technically generalizable to any segmentation task, it is especially useful for tasks where existing algorithms yield initial results with sufficient quality to kickstart the training process, which is the case in tree instance segmentation.

In summary our contributions are:

\begin{itemize}
    \item We propose employing a rating model as ``a weak supervision'' in training segmentation models. Our classifier obtains an accuracy of approximately 90\%, which can be obtained from segmentation quality rating labels established reasonably quickly by a human operator via data inspection.
    \item We validate the proposed method by demonstrating an increase of 34\% in the number of segmented single-tree instances in a challenging alpine dataset, and an important decrease in the proportion of instances that do not correspond to a real tree.
    \item We evaluate the performance of the joint segmentation and rating model against a hand-labeled subset of the dataset. We confirm that our method successfully segments between 80\% to 90\% of trees present in dense forests in the study area, with major performance improvements compared to the state of the art.
\end{itemize}

\section{Related work}
\label{sec:related_work}

\subsection{Point cloud classification}

Point classification models can be broadly divided into two categories. The first category transforms the data into a regular structure, such as 2D image grids \cite{isprs-annals-IV-2-231-2018}, 3D voxel representations \cite{7353481} or slices \cite{Huang_2018_CVPR}, before further processing. Regular structures can be processed by classical architectures such as 2D CNN \cite{isprs-annals-IV-2-231-2018}, 3D CNN \cite{7353481}, RNN \cite{Huang_2018_CVPR}, etc. While CNN have proven easy to use and reliable for image classification tasks, their application on 3D point-clouds is not without an additional inconvenience. Indeed, in 3D data, CNN either impose using a voxel grid, which can be memory intensive in fine resolution or cause information loss in coarse resolution; or projecting 3D data onto a 2D grid, which further removes information from the 3D structure \cite{Qi_2017_CVPR}. Nevertheless, these limitations can be somewhat addressed through the use of Sparse Voxel grids, which have both linear processing and memory complexity \cite{Chen_2023_CVPR}.

An alternative approach is to process lidar point clouds as an irregular data structure. Pioneering this idea was the PointNet architecture, which processes each point individually and uses an ordering invariant function, e.g. maximal or average pooling, to extract features \cite{Qi_2017_CVPR}. A major drawbacks of this approach is the difficulty to capture local detail, while focusing on important structural points \cite{Qi_2017_CVPR}. To partially mitigate this undesirable effect, PointNet++ employs a nested local partitioning of its input \cite{NIPS2017_d8bf84be}. Self-Attention offers another efficient approach to process unstructured collections of objects, which is implemented for point clouds in architectures such as the Point Transformer \cite{Zhao_2021_ICCV}.

When choosing a classification architecture for the proposed learning system, different criteria are of importance, including but not limited to, accuracy, ease of training, and computational speed. We will evaluate three architectures: (i) 3DmFV \cite{8394990}, a dense Voxel Net that mitigates the lower voxel grid resolution by encoding rich local features using 3D Modified Fisher Vectors, (ii) Point Transformer \cite{Zhao_2021_ICCV}, as a state of the art unstructured model, and (iii) a larger resolution 3D CNN we derived from VoxNet \cite{7353481}.

\subsection{Individual tree segmentation}

Closed form point cloud segmentation solutions such as watershed \cite{9033973}, graph cut \cite{7500049} or region growing \cite{rs12071078} seem to be gradually phased out in favor of more high-performance deep learning models. 
Nevertheless, they still offer a reasonable level of accuracy and therefore remain popular as baseline models, partly also due to their general applicability for the task of Individual Tree Detection (ITD) and Individual Tree Crown segmentation (ITC). 

Different deep learning architectures have been adapted to the ITD and/or ITC tasks, including PointNet \cite{f12020131}, PointNet++ \cite{f14071327} and 2D CNN version like YOLOv5 \cite{STRAKER2023100045} or Mask R-CNN \cite{isprs-annals-X-1-2024-67-2024} (for the latter, the point cloud data is projected onto a raster depth map) or 3D UNET \cite{isprs-annals-X-1-W1-2023-605-2023}.

Despite the increase of computational speed, the processing of very large outdoor datasets for ITD and ITCs remains challenging. To address this issue, Xiang \etal \cite{isprs-annals-X-1-W1-2023-605-2023} proposed an architecture using Bottom-up instance grouping. Unlike Top-down instance detection, where objects have to be detected first before getting segmented and where tiling could split objects apart, the Bottom-up instance grouping assigns a feature vector to each point within a cloud which are later segmented using an unsupervised method. A major drawback of Bottom-up instance grouping is that the unsupervised segmentation step can be quite unreliable \cite{Jiang_2020_CVPR}. To alleviate this, the model is based on PointGroup, where multiple clustering variants are used to produce multiple candidate clusters. These are later filtered out using a ScoreNet and the overall process increases the reliability of the segmentation step.

The segmentation approach by Xiang \etal was later combined with geometric filtering and data augmentation tailored to forests to create ForAINet \cite{ForAINet2024}. This panoptic segmentation model (i.e. a model outputing both individual tree segmentation and semantic segmentation of the point cloud) is specialized for forestry. The model was later augmented with multi-resolution data to increase its generalization ability and became the \textit{SegmentAnyTree} model \cite{wielgosz_segmentanytree_2024}.

The ForestFormer3D model has been proposed by \cite{xiang2025forestformer3dunifiedframeworkendtoend} as a replacement for ForAINet \cite{ForAINet2024} and SegmentAnyTree \cite{wielgosz_segmentanytree_2024}. While all these developments still rely on a 3D UNet for encoding, ForestFormer3D replaces the unsupervised segmentation step with a transformer-based layer, making the model fully trainable end-to-end. However, at the time of writing, its code base is not yet publicly available and the method has not been peer reviewed; as such is not evaluated in this work.  

Although the method detailed in Sec.~\ref{sec:methods} can be adapted to employ other (or any) backbone segmentation architectures, we have selected \textit{SegmentAnyTree} \cite{wielgosz_segmentanytree_2024} as the state of the art. This is due to it being the most recent peer-reviewed method and due to its training via data augmentation which should make it generalizable to datasets with varying resolution. Both aspects are interesting for the comparison in Sec.~\ref{sec:results} and following conclusions (Sec.~\ref{sec:conclusion}). 




\section{Methods}
\label{sec:methods}

\begin{figure}
    \centering
    \includegraphics[width=\linewidth]{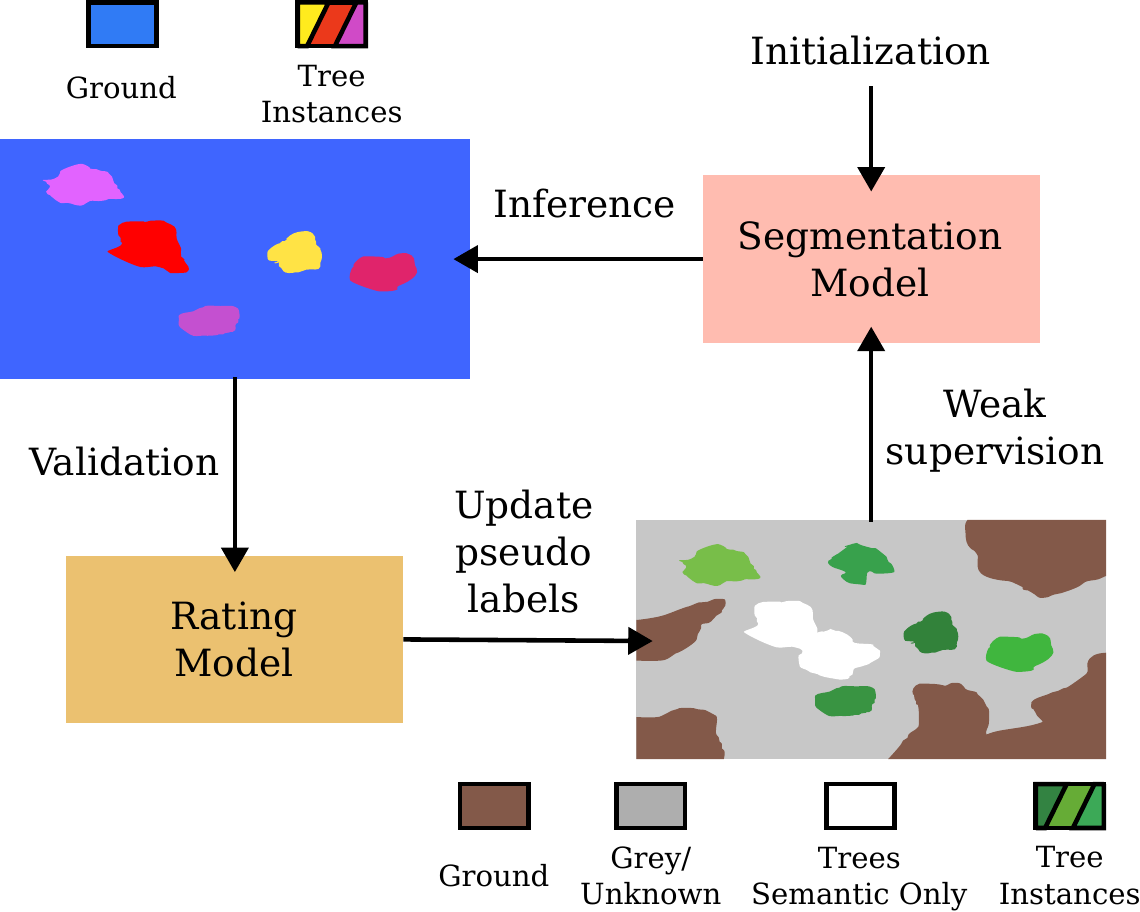}
    \caption{Training loop: knowledge is transferred from the rating model to the segmentation model via an interactive process where the output from the segmentation model is rated and used to build pseudo-labels.}
    \label{fig:training_loop}
\end{figure}

Rating the output of a segmentation algorithm is an order of magnitude faster than manually marking the exact outlines of (tree) clusters. But how can the information encoded in the rating be used to maximally improve the segmentation task? We propose achieving this through an iterative process shown in Figure~\ref{fig:training_loop} and described as follows:

\begin{enumerate}
    \item First, we build an initial segmentation of the data, using either a pre-trained version of the segmentation model (as long as its performance is of sufficient accuracy)
    or a closed form solution such as, \eg watershed \cite{9033973}.
    \item A human operator will then rate a representative sample of the initial clusters obtained in Step 1. For the task of ITCs, the proposed clusters are rated as belonging to one of three pseudo-label tree classes (\className{Single}, \className{Multi} or \className{Non-Tree}),  Sec.~\ref{sec:methods:rating}.
    \item Then, a rating model is trained, \ie a classification model predicting the rating of a cluster, based on the ``ground-truth'' ratings annotated by the human operator (one of the rating models in Sec.~\ref{sec:methods:rating}).
    \item The rating model is subsequently used to rate all clusters predicted in the initial segmentation output. From there, pseudo-label maps are built.
    \item The previously obtained pseudo-labels are used to train the segmentation model (Sec.~\ref{sec:methods:self_superv_seg_mod}).
    \item The updated segmentation predictions are fed to the rating model, the output of which is used to update the pseudo-labels (Sec.~\ref{sec:methods:updt_pseudo_lbl}). From here, we iterate back to step 5, until the number of newly identified tree instances stabilizes.
\end{enumerate}

The rating tool is described in more detail within the supplementary material. The rest of the section will focus on the rating and segmentation models.

\subsection{Supervised rating model}
\label{sec:methods:rating}

To simplify the rating process, instead of attributing a score (which would be somewhat arbitrary and would require the operator to take more time to think), we reduce the rating to a classification problem. Each cluster is assigned to a class which is either \className{Single}, when they are a single tree, \className{Multi}, when multiple trees are captured in the same cluster, or \className{Non-tree}, when other non-tree elements are clustered, or at least the trees are not the dominant elements in the clusters. The human operator in charge of the initial rating only needs a few seconds to make a decision for most clusters.

As for the automated rating model, as stated before, three different architectures were evaluated for the rating model. 3DmFV \cite{8394990} and Point Transformer \cite{Zhao_2021_ICCV} architectures were used "as is", with only the final layer resized to match the three rating classes.

\begin{figure}
    \centering
    \includegraphics[width=\linewidth]{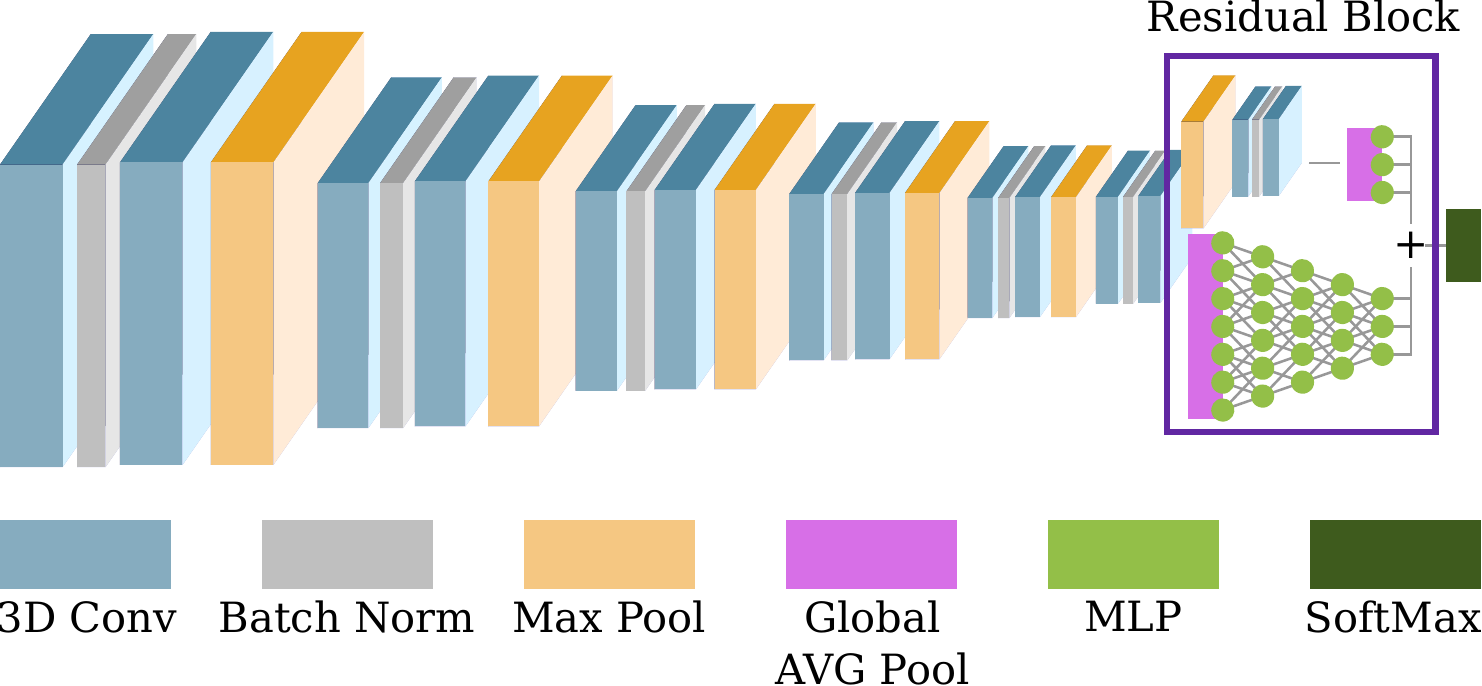}
    \caption{Architecture of proposed rating model}
    \label{fig:voxnetrater}
\end{figure}

We also propose a VoxNet \cite{7353481} architecture, shown in Figure~\ref{fig:voxnetrater}, which trades the rich features of Fisher Vectors used by 3DmFV \cite{8394990} for a higher resolution. Instead of a simple occupancy function \cite{7353481}, we build our voxel grid using Kernel Density Estimation (KDE), which, at the cost of a slightly less sparse voxel grid, provides a much more precise estimate of the local point density, which is in itself an important feature that can be used by the classifier \cite{Hu_2022_CVPR} (\eg to account for the distance between the points and the sensor, which will influence the point density). 

Given a point cloud $\matr{P}$, KDE computes the value of a voxel $\vec{i}$ in the voxel grid $\matr{V}$ as:

\begin{equation}
    \matr{V}[\vec{i}] = \sum_{\vec{p} \in \matr{P}} k(\vec{p} - \vec{i}),
\end{equation}

\noindent with $k$ the kernel function. We used:

\begin{equation}
    k(\vec{x}) = \dfrac{1}{\sqrt{(2 \pi)^3}} \exp{\left( -\dfrac{1}{2} \dotprod{\vec{x}}{\vec{x}} \right) },
\end{equation}

\noindent which is a Gaussian kernel with unit variance.

\begin{figure*}
    \centering
    \includegraphics[width=\textwidth]{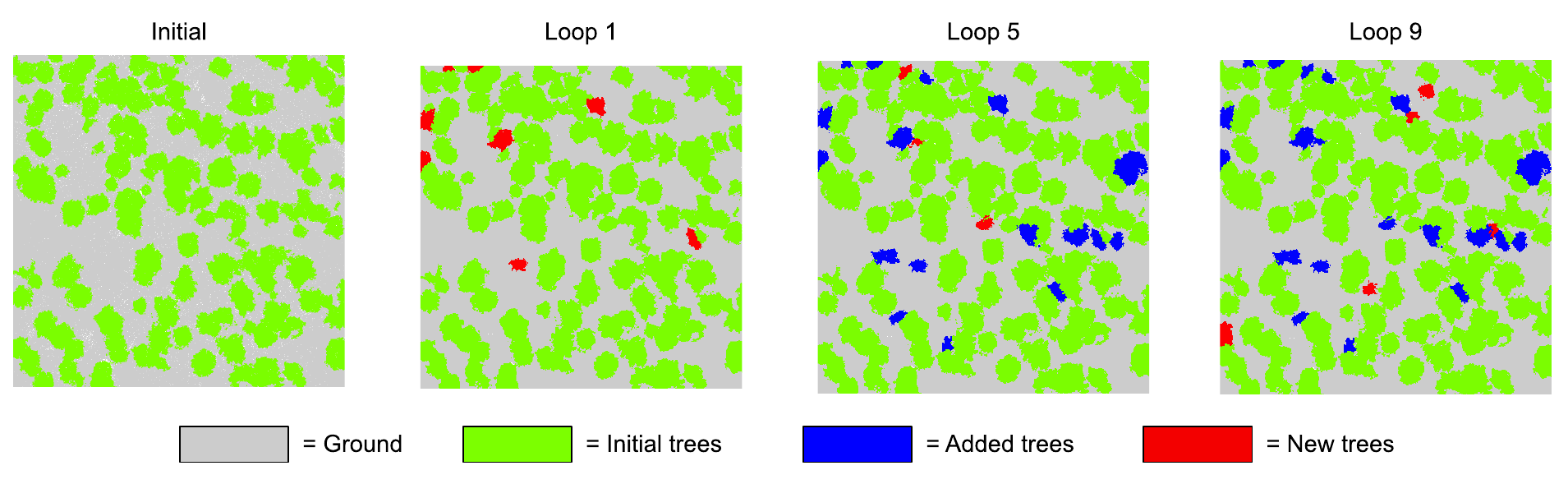}
    \caption{Iterative detection of new trees}
    \label{fig:iterative_tree_detection}
\end{figure*}

The voxel grid is then processed by a sequence of 3D convolutions and Max-pooling layers, halving the resolution but increasing the number of features. The last layers at $1/32$ resolution aggregates all the features using a residual block. The first connection of the residual block uses two convolution layers to reduce the voxel grid to 3 features at $1/64$ resolution which are aggregated using global average pooling. The second connection of the residual block aggregates 1024 features at $1/32$ resolution directly using global average pooling, before processing these features using a multi-layer perceptron. The features from the two branches are summed, and final activations are obtained via a Softmax layer.

Training is done using the ADAM optimizer. We use Batch normalization and weight decay for normalization. For all three models, we also used data augmentation to limit the risk of overfitting by adding random rotations along the $z$ axis. The classification loss is the Cross Entropy Loss, weighted such that the total weights of each class are equivalent (as some classes, especially \className{Non-tree}, can be overrepresented in the initial data), \ie the weights $w$ are computed such that:

\begin{equation}
    w_i c_i = w_j c_j,
\end{equation}

\noindent for any classes $i$ and $j$, where $c_i$ is the number of samples in class $i$.

\subsection{Self-supervised segmentation model}
\label{sec:methods:self_superv_seg_mod}

While all of our experiments will focus on the state of the art \textit{SegmentAnyTree} \cite{wielgosz_segmentanytree_2024} method, the methodology we present here should be adaptable to any method which performs panoptic segmentation of trees, so long as it accepts a binary mask for semantic segmentation, instance mask for tree instances and does not require the training data to be fully labeled (for most loss functions, this can be implemented by setting a weight of 0 to all areas without information).

Pseudo labels are built from the tree clusters as such:

\begin{enumerate}
    \item First, all points in the point cloud are marked as being Ground.
    \item Then, all clusters or points that have been recognized as trees are marked as Gray/Unknown.
    \item Finally, instances of single trees are marked on the data.
\end{enumerate}

Gray/Unknown areas are assigned a weight of $0$ in the loss, while Ground and Trees are processed as usual by the network (\textit{SegmentAnyTree} supports the Gray/Unknown label by default, albeit any method should be adaptable by just setting a weight to the loss, which is supported by all usual loss functions).

The pseudo labels are used for training the model for a number of epochs $n$, before being updated. We performed a simple grid search to tune $n$ and in our experiments we used $n=3$. Albeit the optimal value will vary depending on the size of the dataset and other external factors and will need to be re-estimated in different settings.

\subsection{Update to the pseudo labels}
\label{sec:methods:updt_pseudo_lbl}

After one training loop (step 5), the point cloud is processed with the current parameters of the segmentation model, and the clusters are classified using the rating model. Clusters classified as single trees are then tested to be added to the pseudo labels.

If the candidate single tree does not intersect with any other tree (i.e., it comprises only points classified as Ground or Gray/Unknown), then it is added to the pseudo labels map as a newly identified single tree.

In the case where the candidate overlaps with existing tree points, a series of tests is applied to determine whether the cluster corresponds to a new tree or an already known instance. In particular, to be considered as a new tree, the clusters are evaluated to conform to the following criteria:

\begin{enumerate}
    \item The coordinates of the highest point in the candidate cluster, assumed to be the tip of the tree, should be different than the highest point of all intersecting clusters, as it is assumed that two trees cannot have the same tip.
    \item The tree cannot intersect with other trees at a distance larger than a fixed threshold. We fix the threshold at 2m, as 25\% of the expected width of a tree in the training set which was estimated to be 8m.  
    \item The intersection over cluster  (IoC), for each cluster, is less than 0.7, with
    \begin{equation}
        \text{IoC} = \frac{\text{ \# points of intersection}}{\text{\# points of cluster}}.
    \end{equation}
    \noindent the intersection over cluster is used instead of the intersection over union to avoid small trees intersecting with large trees to be filtered out.
\end{enumerate}

Once a cluster has passed all tests, it is added to the pseudo labels. Points in the intersection of clusters are split between the old and new clusters based on their distance with respect to the centroid of each cluster. Over time, more and more trees will be detected and added to the pseudo labels, as shown in Figure~\ref{fig:iterative_tree_detection}, until no more trees are detected.

\section{Experiments}
\label{sec:results}

To validate the proposed procedure, we test it on a new and somewhat atypical remote sensing dataset, as it would be applied in practice. This section presents the dataset used and experimental evaluation of both the rating model and segmentation model.

\subsection{Dataset}

\begin{figure}
    \centering
    \includegraphics[width=\linewidth]{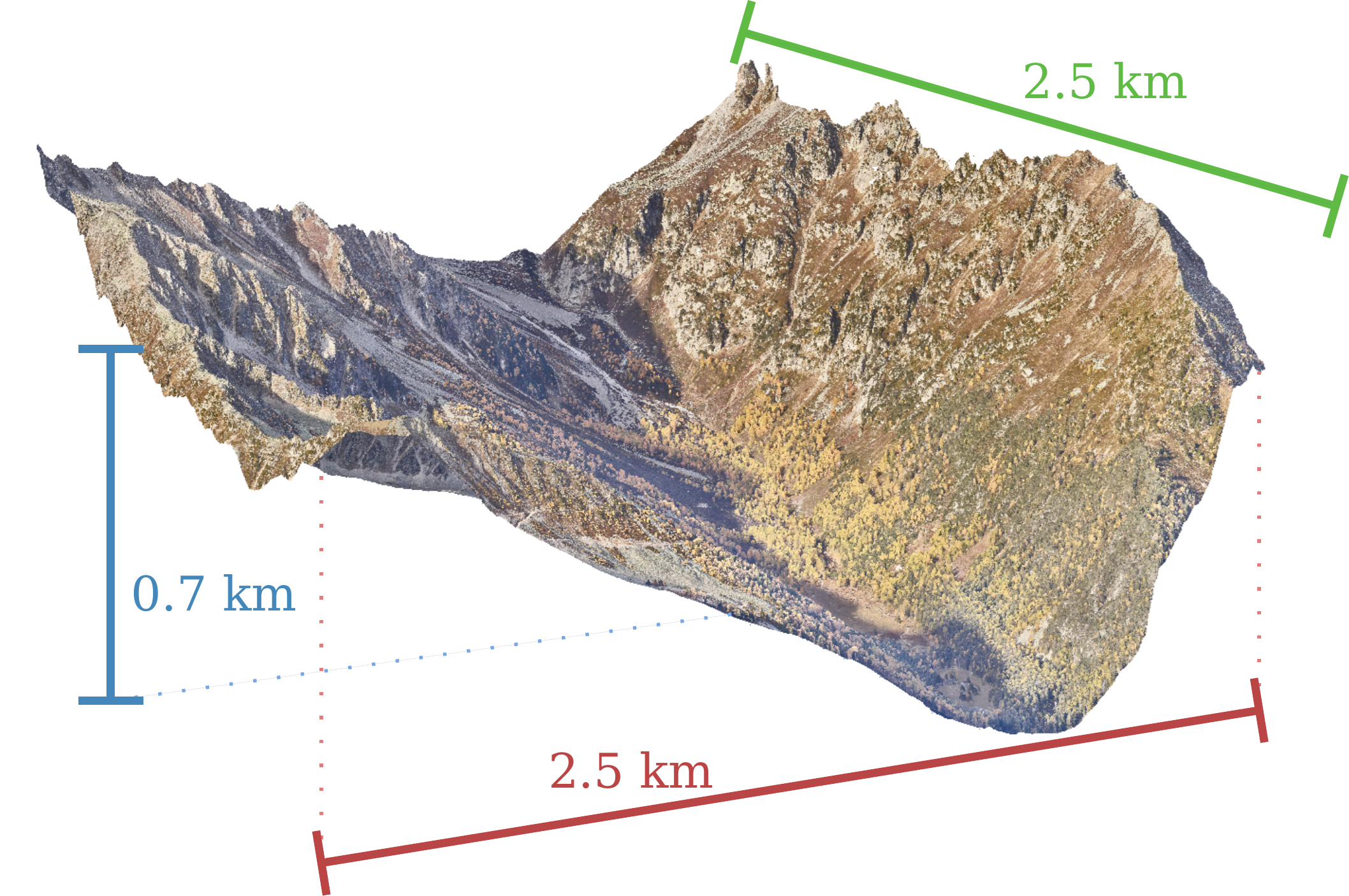}
    \caption{Colorized Lidar point cloud of the study site in the Val d'Arpette region, Switzerland.}
    \label{fig:dataset}
\end{figure}

\begin{figure*}
    \centering
    \includegraphics[width=\linewidth]{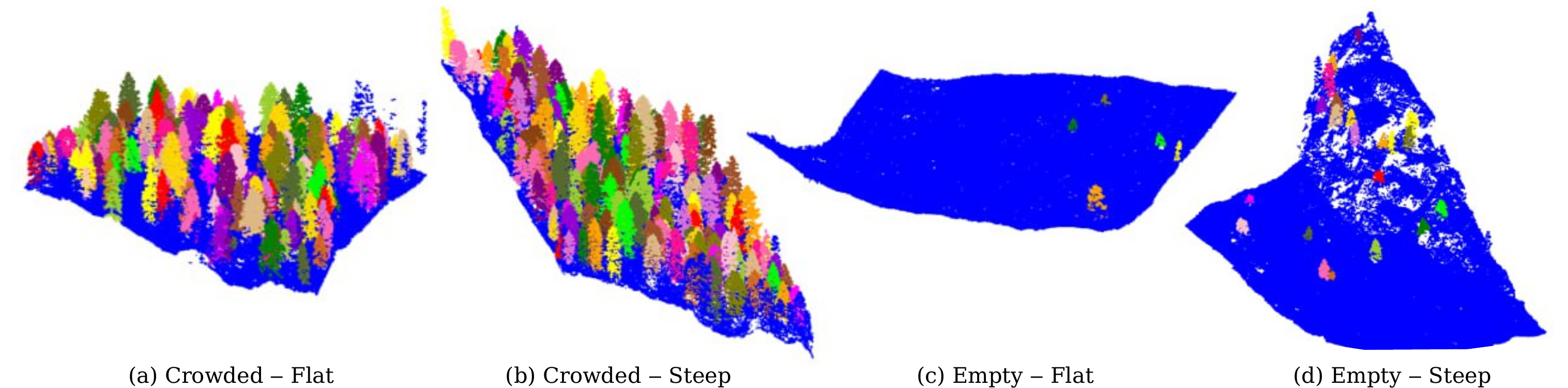}
    \caption{Labeled ground truth tiles.}
    \label{fig:gt_titles}
\end{figure*}

The dataset we will use for this study is a lidar point cloud acquired by an aircraft over a small alpine region in the western part of Switzerland; see Figure~\ref{fig:dataset}, using a state-of-the-art dual channel ALS at the lowest-possible flying height over the mountain ridges to maximize point density. The data was acquired alongside a medium-format aerial camera, which was used to colorize the point cloud. 

The area covers approximately $6.5\,km^2$ with vertical differences up to $\approx700\, m$, includes 272 million points, and has an average density of $38\, pt/m^2$. The total number of trees in this small valley is estimated at around $30'000$ to $40'000$. For comparison, the FOR-instance aerial dataset used for training \textit{SegmentAnyTree} has a point density ranging from around $500\, pt/m^2$ up to $10'000\, pt/m^2$, contained only $0.002\, km^2$ of labeled data, and $1'130$ labeled trees (a ground based dataset was also used for the initial training of \textit{SegmentAnyTree} \cite{rs15153737}, with up to $20'000\, pt/m^2$).

Our point cloud poses multiple challenges for tree segmentation methods (either closed form or deep learning) due to the following reasons:

\begin{itemize}
    \item The very rugged terrain with steep slopes can negatively impact the segmentation models. In certain cases, rocks can be wrongly classified as trees, or slope portions can be included in segmented tree instance clusters.
    \item The time of day and shading, which makes segmentation based on color more challenging.
    \item The mean point density is significantly lower in terms of what would be possible to achieve over a flat terrain from the same ALS (almost by an order of magnitude), and even lower compared to point densities obtained by drone or ground based measurements. It is also on the low end of the synthetic density used for training \textit{SegmentAnyTree}, the lowest density being $10\, pt/m^2$ and second lowest $100\, pt/m^2$.
    \item The point cloud data was collected from an airplane flying at a constant altitude over steep terrain with high alpine ridges on its side. As a result, the distance between the LiDAR sensor and the ground varied, leading to variable point density in the data and adding complexity to the model training process. 
\end{itemize}

An initial segmentation of the full dataset was performed using a commercial implementation of the Watershed algorithm \cite{9033973}, yielding $38'609$ clusters. Of these clusters, $13'316$ were manually rated as representing one of the three classes described in Section~\ref{sec:methods:rating}. Table~\ref{tab:initialClasses} reports the proportion of the different classes for the hand-classified data.

\begin{table}[]
    \centering
    \begin{tabular}{l|rr}
        \className{Single} & $3'790$ & $28.88\%$ \\
        \className{Multi} & $1'448$ & $10.31\%$  \\
        \className{Non-tree} & $7'985$ & $60.81\%$
    \end{tabular}
    \caption{Initial hand classified clusters at Val d'Arpette test.}
    \label{tab:initialClasses}
\end{table}

Due to the computational resources required to train the segmentation model, the dataset was tiled. The size of the tile was set at $100\,m\times100\,m$, a compromise between the size of the tile, ensuring that it contains enough trees, and the processing power of the GPU used for training. The selected tiles were split into a train set, for our method to be executed on, and a test set, reserved for evaluation purposes.

We manually labeled all tree instances of four titles from the test set. To study the behavior of the proposes pipeline in different conditions, we selected a tile in a flat and crowded region, a flat and steep region, an empty and flat region and an empty and steep region; see Figure~\ref{fig:gt_titles}. The ground truth labels are a mix of manually selected tree instances and instances that were missed by the human operator but automatically detected and later validated visually, showing the limit of hand labeling tree clusters. Indeed, manually segmenting trees is not only time consuming, it is also error prone. The major source of error we observed during the labeling process is related to trees that are very close together and that the human operator labeled as a single tree.

\subsection{Classification accuracy}

For each of the three models under consideration, a grid search was performed to tune the most important hyper-parameters of the model. The accuracy and weighted accuracy (\ie accuracy of the model assuming each class having the same weight) are reported in Table~\ref{tab:classifier_results}. The KDE base model we developed reached the highest accuracy ($\sim90\%$) that is, however, closely followed by the Point Transformer architecture. One of the main reasons is that local texture is very important when detecting trees in lidar data. Indeed, airborne laser scanners often operate in frequencies that allow their laser to traverse the upper layer of vegetation and detect deeper branches and leaves, and sometimes even the ground below the vegetation. The 3DmFV \cite{8394990} model struggles to capture this due to its lower resolution, and the Point Transformer architecture, as an architecture optimized for irregular data structure processing, also struggles a little bit.

\begin{table}[]
    \centering
    \resizebox{\linewidth}{!}{
    \begin{tabular}{l|rr}
         & \textbf{Accuracy} & \textbf{Weighted Accuracy} \\
         \hline
        3DmFV \cite{8394990} &  $83.4\%$ & $73.0\%$ \\
        Point Transformer \cite{Zhao_2021_ICCV} & $87.3\%$ & $85.5\%$ \\
        KDE (ours) & $\mathbf{91.6\%}$ & $\mathbf{88.6\%}$
    \end{tabular}
    }
    \caption{Accuracy of the 3 studied classifier models}
    \label{tab:classifier_results}
\end{table}

Note that the performance of all models decreases when each class is re-weighted. Despite the weights in the Cross Entropy loss, the models did overfit the \className{Non-tree} class slightly, as this class is overrepresented in the training data. Nevertheless, the effect is minimal, so we opted to proceed with the KDE VoxNet based classifier as our rating model.

\subsection{Single tree segmentation}

To evaluate the proposed iterative segmentation approach, we will use the classification rating model to estimate the quality of the segmentation. This is to check if the number of segmented single trees increases per iteration, as well as if the number of multi-tree and non-tree clusters decreases. We re-trained the \textit{SegmentAnyTree} model on our data using the pseudo labels established by the rating model at a learning rate of $5 \cdot 10{-5}$, which was found to provide the best compromise between speed and stability of training.

\begin{figure}
    \centering
    \includegraphics[width=\linewidth]{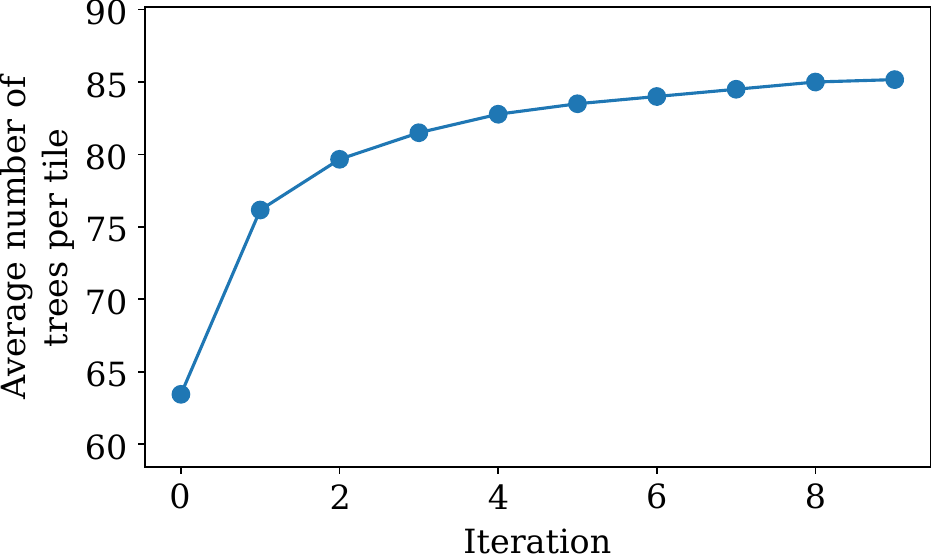}
    \caption{Average trees per tile at each iteration of the pipeline}
    \label{fig:avg_tree_tile}
\end{figure}

\begin{figure}
    \centering
    \includegraphics[width=\linewidth]{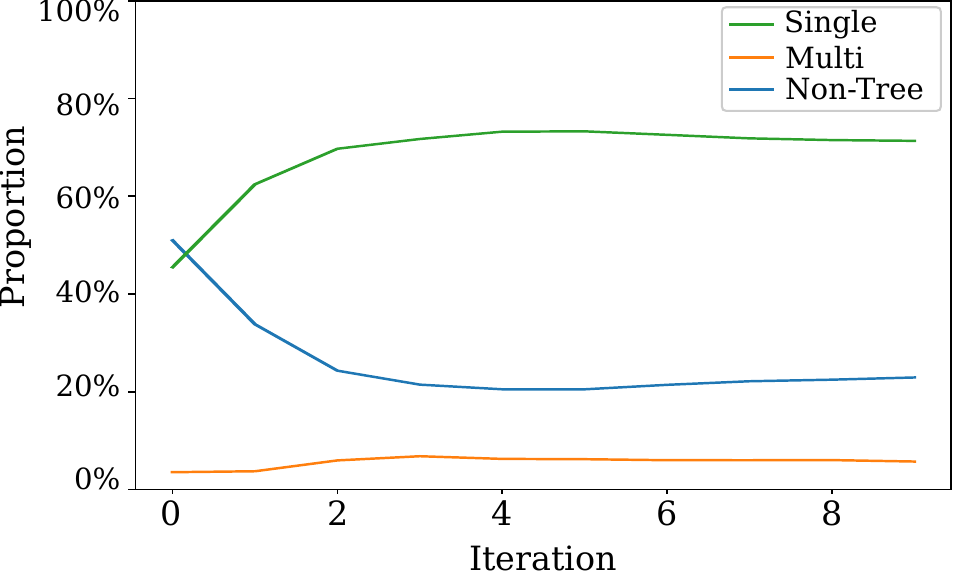}
    \caption{Proportion of the different rating classes predicted by the rating model as the training progresses}
    \label{fig:rating_progress}
\end{figure}

As shown in Figure~\ref{fig:avg_tree_tile}, the average number of identified trees per tile (surface of one hectare) increased from 63.4 at iteration 0 to 85.2 at iteration 9, an increase of 34.3 trees per hectare. The increase in the number of detected trees is very sharp at the beginning and stabilizes over time. Figure~\ref{fig:rating_progress} also shows that the proportion of \className{Non-tree} samples goes down as training progresses, while the proportion \className{Multi} clusters remains more or less constant. This indicates that the  proposed training scheme increases the number of retrieved \className{Single} tree instances and reduces the number of clusters assigned as \className{Non-tree}.

\subsection{Segmentation with respect to manual ground truth}

While the increase in detected trees by the segmentation model is encouraging, it is also important to compare it with the actual number of trees on the ground.
For that we employ the manually labeled ground truth depicted in Figure~\ref{fig:gt_titles}. 

The comparison is the following. We take the output of either the original \textit{SegmentAnyTree} model, or its evolved version by our approach, and process it with the described rating model so that non-\className{Single} tree clusters are filtered out. This shows another benefit of our approach: the rating model can be combined with the segmentation model at inference time to refine its output. We finally compare the number of predicted trees in the cluster against the ground truth.

The results, reported in Table~\ref{tab:segmentation_results} per terrain steepness and tree population density, show that while our method does not reach a coverage of 100\%, at least with 9 iterations, it increases the number of detected instances. The performance is not as strong for sparsely populated tiles containing few isolated trees. This indicates that the method is challenged by the presence of small shrubs that are slightly out of domain compared to the typical tree in the dataset. 

\begin{table}[]
    \centering
    \begin{tabular}{l|r}
         & \textbf{\# instances} \\
         \hline
         \multicolumn{2}{l}{\textbf{(a) Crowded - flat}} \\
        SegmentAnyTree Original &  $136$ \\
        SegmentAnyTree + Ours & $\mathbf{162}$ \\
        Ground truth & $\textcolor{gray}{182}$ \\
         \multicolumn{2}{l}{\textbf{(b) Crowded - steep}} \\
        SegmentAnyTree Original &  $162$ \\
        SegmentAnyTree + Ours & $\mathbf{189}$ \\
        Ground truth & $\textcolor{gray}{200}$ \\
         \multicolumn{2}{l}{\textbf{(c) Empty - flat}} \\
        SegmentAnyTree Original &  $2$ \\
        SegmentAnyTree + Ours & $2$ \\
        Ground truth & $\textcolor{gray}{5}$ \\
         \multicolumn{2}{l}{\textbf{(d) Empty - steep}} \\
        SegmentAnyTree Original &  $10$ \\
        SegmentAnyTree + Ours & $\mathbf{15}$ \\
        Ground truth & $\textcolor{gray}{22}$
    \end{tabular}
    \caption{Instances detected with respect to ground truth for labeled ground truth}
    \label{tab:segmentation_results}
\end{table}

\begin{figure*}
    \includegraphics[width=\linewidth]{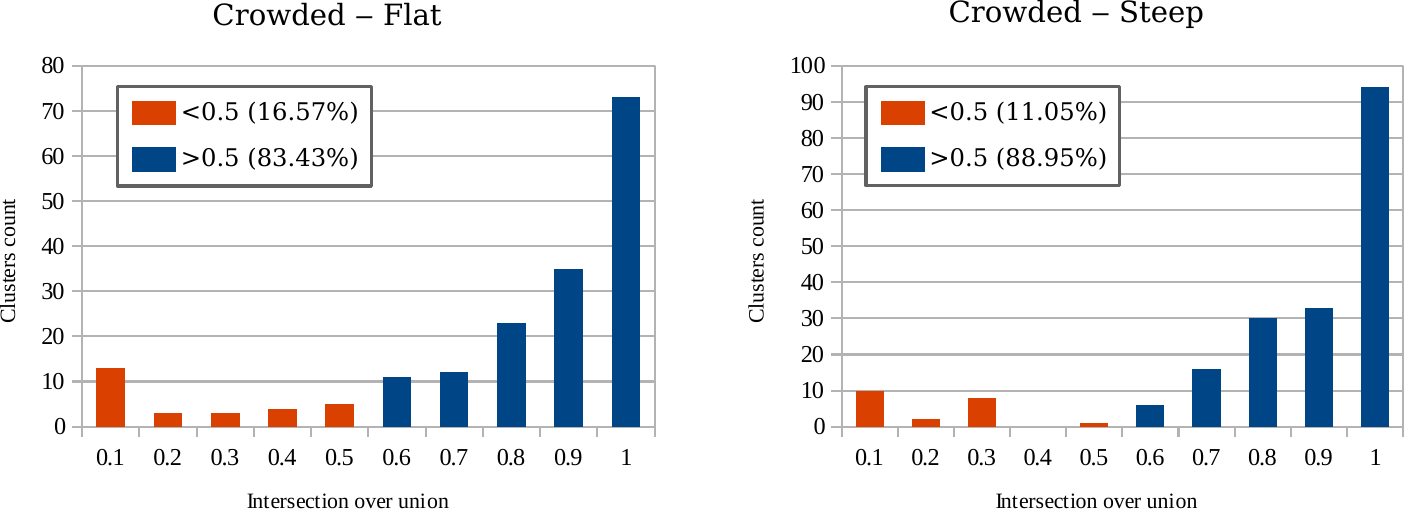}
    \caption{IOU between ground truth clusters and corresponding predictions of the finetuned model on ground truth crowded tiles}
    \label{fig:iou_gt}
    
\end{figure*}

A more detailed analysis of the results shown in Figure~\ref{fig:iou_gt} is in very good agreement with the ground truth for densely populated areas independent of slope steepness. Only a certain number of trees from the ground truth were missed, leading to very small IOU. For crowded tiles more than 80\% of ground truth clusters have an IOU of more than 0.5 with their corresponding predicted clusters. The results are more concerning for ``empty'' tiles (discussed in the supplementary material), where not only were most of the present small trees not detected, but also some proportion of the segmented trees do not match the ground truth.

\section{Conclusion}
\label{sec:conclusion}

We proposed a method which enables employing efficiently obtained rating labels by a human operator to facilitate instance segmentation of trees in lidar point cloud data obtained in a challenging  environment. The proposed method resulted in a significant increase in the proportion of single trees detected by a state-of-the-art deep learning model in alpine terrain close to a tree limit and, when combined with our automatic rating model, enabled the recovery of more than $80\%$ of the trees in the dense validation tiles that were annotated by hand.
Yet, challenges remain in detecting small and isolated trees growing in rough terrain. 
This is partially due to the fact that the trees in these regions are very small and out of distribution compared to the main population of trees in the forest.

Our results demonstrate the potential benefits of using a simple rating model for weak supervision within instance segmentation of tree individuals.
The rating model can also be used in conjunction with the segmentation model at inference time to filter out invalid clusters. This provides additional value to the segmentation model.

\section{Acknowledgments}
\label{sec:acknowledgments}

Swann Emilien Céleste Destouches conducted the experiments described in this paper under the supervision of Jesse Lahaye and Laurent Jospin. Experimental design was done by Laurent Jospin with additional contributions by Swann Emilien Céleste Destouches. The tool for rating point clusters was developed by Laurent Jospin. The tool used for hand labeling the clusters in the point cloud data was developed by Swann Emilien Céleste Destouches. The overall research was directed by Jan Skaloud. Thanks to Swiss Flying Services and SixSense Helimap SA for facilitating the acquisition and preprocessing of our dataset.

{
\small
\bibliography{main}

\begin{thebibliography}{xx}

\bibitem[Ben-Shabat et al., 2018]{8394990}
Ben-Shabat, Y., Lindenbaum, M., Fischer, A., 2018.
 3DmFV: Three-Dimensional Point Cloud Classification in Real-Time Using
  Convolutional Neural Networks.
 {\em IEEE Robotics and Automation Letters}, 3(4), 3145-3152.

\bibitem[Chaudhari et al., 2025]{10.1145/3743127}
Chaudhari, S., Aggarwal, P., Murahari, V., Rajpurohit, T., Kalyan, A.,
  Narasimhan, K., Deshpande, A., Castro~da Silva, B., 2025.
 RLHF Deciphered: A Critical Analysis of Reinforcement Learning from Human
  Feedback for LLMs.
 {\em ACM Comput. Surv.}

\bibitem[Chen et al., 2021]{f12020131}
Chen, X., Jiang, K., Zhu, Y., Wang, X., Yun, T., 2021.
 Individual Tree Crown Segmentation Directly from UAV-Borne LiDAR Data Using
  the PointNet of Deep Learning.
 {\em Forests}, 12(2).
 https://www.mdpi.com/1999-4907/12/2/131.

\bibitem[Chen et al., 2023]{Chen_2023_CVPR}
Chen, Y., Liu, J., Zhang, X., Qi, X., Jia, J., 2023.
 Voxelnext: Fully sparse voxelnet for 3d object detection and tracking.
 \emph{Proceedings of the IEEE/CVF Conference on Computer Vision and Pattern
  Recognition (CVPR)}, 21674--21683.

\bibitem[Cheng et al., 2023]{Cheng_2023_CVPR}
Cheng, T., Wang, X., Chen, S., Zhang, Q., Liu, W., 2023.
 Boxteacher: Exploring high-quality pseudo labels for weakly supervised
  instance segmentation.
 \emph{Proceedings of the IEEE/CVF Conference on Computer Vision and Pattern
  Recognition (CVPR)}, 3145--3154.

\bibitem[Dalponte and Coomes, 2016]{10.1111_2041-210X.12575}
Dalponte, M., Coomes, D.~A., 2016.
 Tree-centric mapping of forest carbon density from airborne laser scanning and
  hyperspectral data.
 {\em Methods in Ecology and Evolution}, 7(10), 1236-1245.

\bibitem[Fan et al., 2024]{isprs-annals-X-1-2024-67-2024}
Fan, W., Tian, J., Troles, J., D\"ollerer, M., Kindu, M., Knoke, T., 2024.
 Comparing Deep Learning and MCWST Approaches for Individual Tree Crown
  Segmentation.
 {\em ISPRS Annals of the Photogrammetry, Remote Sensing and Spatial
  Information Sciences}, X-1-2024, 67--73.
 https://isprs-annals.copernicus.org/articles/X-1-2024/67/2024/.

\bibitem[Fouqueray et al., 2020]{FOUQUERAY2020117880}
Fouqueray, T., Charpentier, A., Trommetter, M., Frascaria-Lacoste, N., 2020.
 The calm before the storm: How climate change drives forestry evolutions.
 {\em Forest Ecology and Management}, 460, 117880.

\bibitem[Hu et al., 2022]{Hu_2022_CVPR}
Hu, J. S.~K., Kuai, T., Waslander, S.~L., 2022.
 Point density-aware voxels for lidar 3d object detection.
 \emph{Proceedings of the IEEE/CVF Conference on Computer Vision and Pattern
  Recognition (CVPR)}, 8469--8478.

\bibitem[Huang et al., 2018]{Huang_2018_CVPR}
Huang, Q., Wang, W., Neumann, U., 2018.
 Recurrent slice networks for 3d segmentation of point clouds.
 \emph{Proceedings of the IEEE Conference on Computer Vision and Pattern
  Recognition (CVPR)}.

\bibitem[Jiang et al., 2023]{Jiang2023}
Jiang, H., Zou, Q., Zhou, B., Jiang, Y., Cui, J., Yao, H., Zhou, W., 2023.
 Estimation of Shallow Landslide Susceptibility Incorporating the Impacts of
  Vegetation on Slope Stability.
 {\em International Journal of Disaster Risk Science}, 14(4), 618-635.

\bibitem[Jiang et al., 2020]{Jiang_2020_CVPR}
Jiang, L., Zhao, H., Shi, S., Liu, S., Fu, C.-W., Jia, J., 2020.
 Pointgroup: Dual-set point grouping for 3d instance segmentation.
 \emph{Proceedings of the IEEE/CVF Conference on Computer Vision and Pattern
  Recognition (CVPR)}.

\bibitem[Lee et al., 2016]{7500049}
Lee, J., Cai, X., Lellmann, J., Dalponte, M., Malhi, Y., Butt, N., Morecroft,
  M., Schönlieb, C.-B., Coomes, D.~A., 2016.
 Individual Tree Species Classification From Airborne Multisensor Imagery Using
  Robust PCA.
 {\em IEEE Journal of Selected Topics in Applied Earth Observations and Remote
  Sensing}, 9(6), 2554-2567.

\bibitem[Liu et al., 2023]{f14071327}
Liu, Y., You, H., Tang, X., You, Q., Huang, Y., Chen, J., 2023.
 Study on Individual Tree Segmentation of Different Tree Species Using
  Different Segmentation Algorithms Based on 3D UAV Data.
 {\em Forests}, 14(7).

\bibitem[Ma et al., 2020]{rs12071078}
Ma, Z., Pang, Y., Wang, D., Liang, X., Chen, B., Lu, H., Weinacker, H., Koch,
  B., 2020.
 Individual Tree Crown Segmentation of a Larch Plantation Using Airborne Laser
  Scanning Data Based on Region Growing and Canopy Morphology Features.
 {\em Remote Sensing}, 12(7).

\bibitem[Maturana and Scherer, 2015]{7353481}
Maturana, D., Scherer, S., 2015.
 Voxnet: A 3d convolutional neural network for real-time object recognition.
 \emph{2015 IEEE/RSJ International Conference on Intelligent Robots and Systems
  (IROS)}, 922--928.

\bibitem[Qi et al., 2017a]{Qi_2017_CVPR}
Qi, C.~R., Su, H., Mo, K., Guibas, L.~J., 2017a.
 Pointnet: Deep learning on point sets for 3d classification and segmentation.
 \emph{Proceedings of the IEEE Conference on Computer Vision and Pattern
  Recognition (CVPR)}.

\bibitem[Qi et al., 2017b]{NIPS2017_d8bf84be}
Qi, C.~R., Yi, L., Su, H., Guibas, L.~J., 2017b.
 Pointnet++: Deep hierarchical feature learning on point sets in a metric
  space.
 I.~Guyon, U.~V. Luxburg, S.~Bengio, H.~Wallach, R.~Fergus, S.~Vishwanathan,
  R.~Garnett (eds), \emph{Advances in Neural Information Processing Systems},
  ~30, Curran Associates, Inc.

\bibitem[Rizaldy et al., 2018]{isprs-annals-IV-2-231-2018}
Rizaldy, A., Persello, C., Gevaert, C.~M., Oude~Elberink, S.~J., 2018.
 FULLY CONVOLUTIONAL NETWORKS FOR GROUND CLASSIFICATION FROM LIDAR POINT
  CLOUDS.
 {\em ISPRS Annals of the Photogrammetry, Remote Sensing and Spatial
  Information Sciences}, IV-2, 231--238.

\bibitem[Shimoda and Yanai, 2019]{Shimoda_2019_ICCV}
Shimoda, W., Yanai, K., 2019.
 Self-supervised difference detection for weakly-supervised semantic
  segmentation.
 \emph{Proceedings of the IEEE/CVF International Conference on Computer Vision
  (ICCV)}.

\bibitem[Straker et al., 2023]{STRAKER2023100045}
Straker, A., Puliti, S., Breidenbach, J., Kleinn, C., Pearse, G., Astrup, R.,
  Magdon, P., 2023.
 Instance segmentation of individual tree crowns with YOLOv5: A comparison of
  approaches using the ForInstance benchmark LiDAR dataset.
 {\em ISPRS Open Journal of Photogrammetry and Remote Sensing}, 9, 100045.

\bibitem[Wielgosz et al., 2023]{rs15153737}
Wielgosz, M., Puliti, S., Wilkes, P., Astrup, R., 2023.
 Point2Tree(P2T)—Framework for Parameter Tuning of Semantic and Instance
  Segmentation Used with Mobile Laser Scanning Data in Coniferous Forest.
 {\em Remote Sensing}, 15(15).

\bibitem[Wielgosz et al., 2024]{wielgosz_segmentanytree_2024}
Wielgosz, M., Puliti, S., Xiang, B., Schindler, K., Astrup, R., 2024.
 {SegmentAnyTree}: {A} sensor and platform agnostic deep learning model for
  tree segmentation using laser scanning data.
 {\em Remote Sensing of Environment}, 313, 114367.

\bibitem[Xiang et al., 2023]{isprs-annals-X-1-W1-2023-605-2023}
Xiang, B., Peters, T., Kontogianni, T., Vetterli, F., Puliti, S., Astrup, R.,
  Schindler, K., 2023.
 TOWARDS ACCURATE INSTANCE SEGMENTATION IN LARGE-SCALE LIDAR POINT CLOUDS.
 {\em ISPRS Annals of the Photogrammetry, Remote Sensing and Spatial
  Information Sciences}, X-1/W1-2023, 605--612.

\bibitem[Xiang et al., 2024]{ForAINet2024}
Xiang, B., Wielgosz, M., Kontogianni, T., Peters, T., Puliti, S., Astrup, R.,
  Schindler, K., 2024.
 Automated forest inventory: analysis of high-density airborne LiDAR point
  clouds with 3D deep learning.
 {\em Remote Sensing of Environment}, 305, 114078.

\bibitem[Xiang et al., 2025]{xiang2025forestformer3dunifiedframeworkendtoend}
Xiang, B., Wielgosz, M., Puliti, S., Král, K., Krůček, M., Missarov, A.,
  Astrup, R., 2025.
 Forestformer3d: A unified framework for end-to-end segmentation of forest
  lidar 3d point clouds.

\bibitem[Yang et al., 2020]{9033973}
Yang, J., Kang, Z., Cheng, S., Yang, Z., Akwensi, P.~H., 2020.
 An Individual Tree Segmentation Method Based on Watershed Algorithm and
  Three-Dimensional Spatial Distribution Analysis From Airborne LiDAR Point
  Clouds.
 {\em IEEE Journal of Selected Topics in Applied Earth Observations and Remote
  Sensing}, 13, 1055-1067.

\bibitem[Zhao et al., 2021]{Zhao_2021_ICCV}
Zhao, H., Jiang, L., Jia, J., Torr, P.~H., Koltun, V., 2021.
 Point transformer.
 \emph{Proceedings of the IEEE/CVF International Conference on Computer Vision
  (ICCV)}, 16259--16268.

\end{thebibliography}
}

\end{document}